%% file: template.tex
\documentclass{article}

\usepackage{arxiv}
\usepackage{amsmath,amssymb,amsfonts}
\usepackage{algorithmic}
\usepackage{array}
\usepackage[caption=false,font=normalsize,labelfont=sf,textfont=sf]{subfig}
\usepackage{textcomp}
\usepackage{stfloats}
\usepackage{url}
\usepackage{cite}
\usepackage{booktabs}
\usepackage{etoolbox}

\usepackage{colortbl}
\usepackage{verbatim}
\usepackage{graphicx}
\usepackage{adjustbox}
\usepackage{balance}

\usepackage[numbers]{natbib}
\usepackage[framemethod=tikz]{mdframed}
\usepackage[ruled, linesnumbered, lined, boxed, commentsnumbered]{algorithm2e} %
\usepackage{color}

\usepackage{multirow}

\usepackage[switch]{lineno}
\usepackage{threeparttable}
\usepackage[normalem]{ulem} 
\usepackage{tabularx} 
\usepackage{booktabs} 
\usepackage{makecell}
\usepackage{lipsum} 
\usepackage{float} 
\usepackage{enumitem} 
\usepackage{tcolorbox} 

\usepackage{textcomp} 

\newtheorem{definition}{Definition}

\usepackage{hyperref}
\usepackage{graphics}
\usepackage[caption=false]{subfig}
\usepackage[section]{placeins}

\title{Few-shot Image Classification based on Gradual Machine Learning}

\author{{Na Chen\textsuperscript{1}}, {Xianming Kuang\textsuperscript{2}}, {Feiyu Liu\textsuperscript{1}}, {Kehao Wang\textsuperscript{2}}, {Qun Chen\textsuperscript{1,2}}\thanks{Joint corresponding author:chenbenben@nwpu.edu.cn}\\ \\
\textsuperscript{1}School of software,Northwestern Polytechnical University,Xi'an, China, 710072 \\
\textsuperscript{2}School of Computer Science,Northwestern Polytechnical University,Xi'an, China, 710072 \\
}

\renewcommand{\headeright}{Technical Report}
\renewcommand{\undertitle}{Technical Report}
\renewcommand{\shorttitle}{Few-shot Image Classification based on Gradual Machine Learning}

\hypersetup{
pdftitle={Few-shot Image Classification based on Gradual Machine Learning},
pdfsubject={q-bio.NC, q-bio.QM},
pdfauthor={Na Chen, Xianming Kuang, Feiyu Liu, Kehao Wang and Qun Chen},
pdfkeywords={First keyword, Second keyword, More},
}

\begin{document}
\maketitle

\begin{abstract}
	Few-shot image classification aims to accurately classify unlabeled images using only a few labeled samples. The state-of-the-art solutions are built by deep learning, which focuses on designing increasingly complex deep backbones. Unfortunately, the task remains very challenging due to the difficulty of transferring the knowledge learned in training classes to new ones. In this paper, we propose a novel approach based on the non-i.i.d paradigm of gradual machine learning (GML). It begins with only a few labeled observations, and then gradually labels target images in the increasing order of hardness by iterative factor inference in a factor graph. Specifically, our proposed solution extracts indicative feature representations by deep backbones, and then constructs both unary and binary factors based on the extracted features to facilitate gradual learning. The unary factors are constructed based on class center distance in an embedding space, while the binary factors are constructed based on k-nearest neighborhood. We have empirically validated the performance of the proposed approach on benchmark datasets by a comparative study. Our extensive experiments demonstrate that the proposed approach can improve the SOTA performance by 1-5\% in terms of accuracy. More notably, it is more robust than the existing deep models in that its performance can consistently improve as the size of query set increases while the performance of deep models remains essentially flat or even becomes worse.
\end{abstract}


\maketitle
\input{1.introduction.tex}
\input{2.related_work.tex}
\input{3.Task_Statement.tex}

\input{4.The_GML_Framework.tex}
\input{5.Factor_Graph_Construction.tex}
\input{6.Experiments.tex}
\input{7.Conclusion.tex}

\bibliographystyle{unsrtnat}
\bibliography{references}  






\end{document}

%% file: 1.introduction.tex
\section{Introduction}

Extensively studied in the literature, image classification can be accurately performed by Deep Neural Network (DNN) models provided that there are sufficient labeled training data~\cite{schmarje2021survey,2021Fair}. Unfortunately, in many application scenarios (e.g., medical image analysis~\cite{2021Interactive} and autonomous driving~\cite{yu2020rein}), large amounts of labeled images may not be readily available because data acquisition and annotation needs to involve intensive manual effort. Under these circumstances, DNN models can easily overfit and fail to achieve satisfactory performance. To address this limitation, \textit{few-shot image classification} has been proposed to classify unseen classes with only a few labeled samples~\cite{2020Generalizing,2022A}.

The existing approaches for few-shot image classification can be broadly categorized into two groups: \textit{inductive few-shot learning} and \textit{transductive few-shot learning}. Inductive learning typically trains a generic model based on the labeled samples in training classes, and then directly uses the learnt model to classify each unlabeled sample in test classes independently from each other~\cite{0Siamese,2016Matching,2017Prototypical,sung2018learning,ravi2017optimization,2017Model}. In contrast, supposing that it has the access to both labeled and unlabeled samples in test classes,  transductive learning performs class label inference jointly for all the unlabeled samples.
	
\begin{figure*}[htbp]
	\centering
	\includegraphics[width=0.92\textwidth]{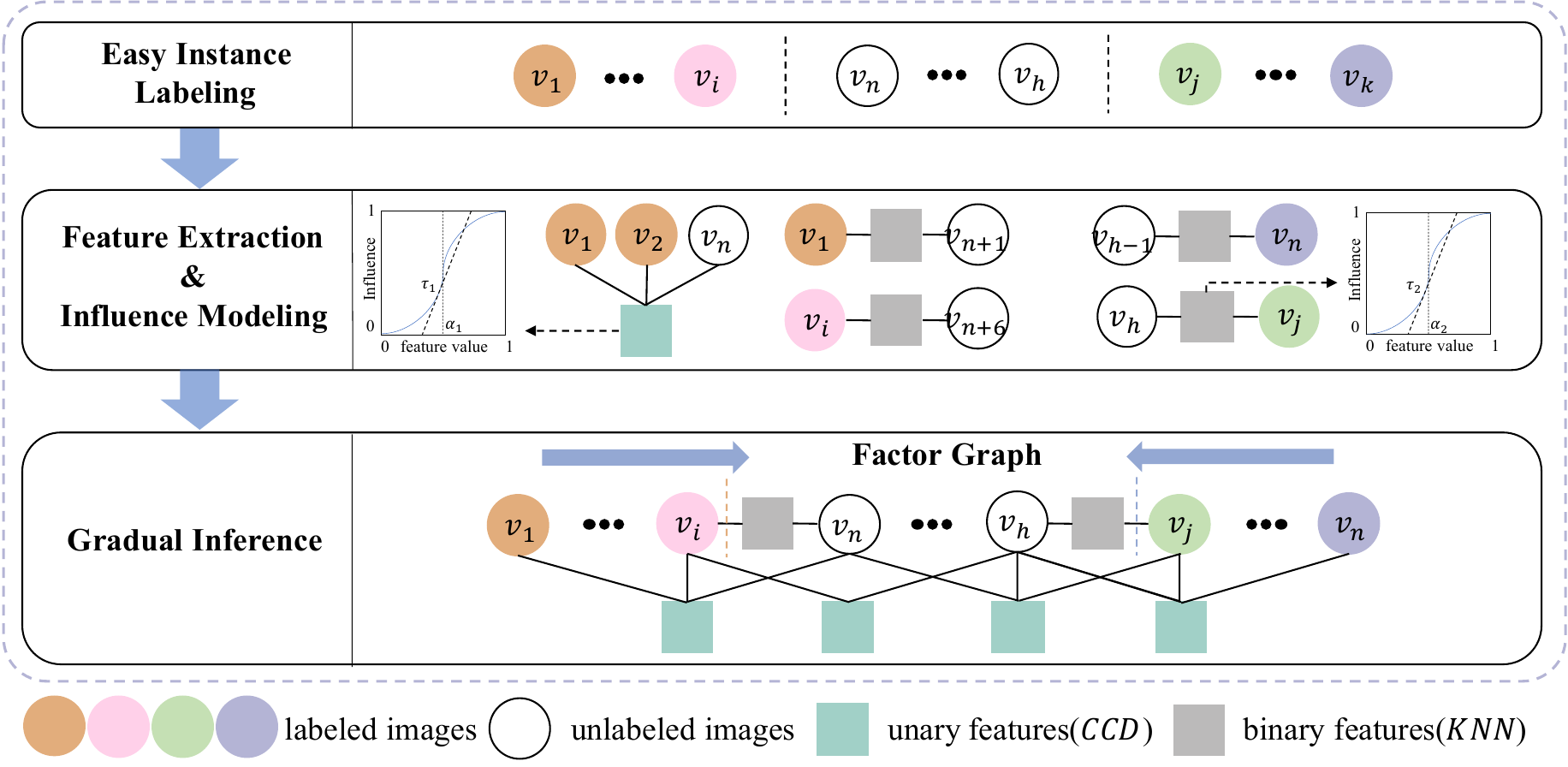}
	\caption{The framework of GML for few-shot image classification: there are two types of factors in the constructed factor graph, which correspond to the unary class centroid distance (\emph{CCD}) and binary k-nearest neighborhood (\emph{KNN}) features. }
	\label{fig:model structure}
\end{figure*}

It can be observed that the core challenge of few-shot image classification is to transfer the knowledge learned on training classes with sufficient labeled samples to new classes with only a few labeled samples. Due to distribution drift between classes, the existing deep learning solutions, whether inductive or transductive, whose efficacy depends on the i.i.d (Independently and Identically Distributed) assumption, have so far achieved only limited success. Compared with inductive learning, transductive learning usually performs better because it can leverage additional sample relationships to fine-tune the distributions of unseen classes. However, it still aims to learn a unified global model for new class prediction. As a result, its efficacy is similarly constrained by scarcity of label observations on new classes.


  To alleviate such limitation, this paper proposes a novel approach for few-shot image classification based on the non-i.i.d paradigm of Gradual Machine Learning (GML). GML was first proposed for the task of entity resolution~\cite{hou2019gradual,HouTKDE,zhong2021attention}, but then also applied to sentiment analysis~\cite{2020Aspect,ahmed2021dnn,wang2023supervised}. Inspired by the nature of human learning, GML begins with some labeled samples, and then gradually classifies unlabeled samples in the increasing order of hardness by iterative knowledge conveyance. Technically, GML fulfills gradual learning by iterative factor inference in a factor graph. It is noteworthy that	instead of learning a unified global classifier, GML gradually classifies one sample at a time by factor inference based on its own evidential certainty.

We have sketched the structure of factor graph for few-shot image classification in Figure.~\ref{fig:model structure}. As shown in the figure, the labeled samples in the support set and the unlabeled samples in the query set are modeled as labeled and unlabeled variables respectively in the factor graph. The labeled samples serve as initial evidential observations, while the unlabeled samples are supposed to be gradually classified by iterative factor inference. In GML, the efficacy of gradual inference depends on effective feature mechanisms for knowledge conveyance. Due to the big advances of deep neural networks for image classification, DNNs are considerably more effective at discriminative feature extraction compared with traditional approaches~\cite{afrasiyabi2022matching,zhao2022self}. Therefore, we leverage the existing deep neural backbones for few-shot image classification to extract discriminative image features, and then model them as factors in a factor graph to facilitate knowledge conveyance. Specifically, we generate vector representations for images in a deep class-sensitive embedding space, and then extract their discriminative features based on the monotonous metrics of class centroid distance and k-nearest neighborhood. Intuitively, the closer to a class centroid an image is, the more likely it belongs to the class. Similarly, the more same-class nearest neighbors an image has, the more likely it belongs to the same class. 
	
	
  To construct diverse mechanisms for knowledge conveyance, our solution uses different deep backbones to generate separate embedding spaces. Our algorithm leverages ResNet-12~\cite{2016Deep} and WRN-28-10~\cite{2016Wide}, both of which are popular backbones for few-shot image classification, for feature extraction. It is noteworthy that even though the WRN-28-10 network is constructed based on ResNet-12, it has considerably wider residual blocks, and thus more advanced representational capabilities. Our experiments have shown that the ResNet-12 and WRN-28-10 networks are to some extent complementary to each other. As a result, leveraging both of them for gradual inference can effectively improve few-shot classification accuracy. 

Since our proposed solution reasons about the labels of test samples in a joint manner, it can be generally considered as an approach of transductive learning. The main contributions of this paper can be summarized as follows:

  \begin{itemize}
  \item We propose a novel approach for few-shot image classification based on the non-i.i.d paradigm of GML, which can gradually classify unlabeled samples in the increasing order of hardness given only a few initial labeled samples. The proposed approach can be potentially generalized to other few-shot classification tasks. 
    
  \item We present a factor graph model consisting of both unary and binary monotonous factors, which can be easily extracted by the existing deep backbones, to enable gradual learning for few-shot image classification. 
  
  \item We empirically validate the performance of the proposed approach on real benchmark datasets by a comparative study. Our extensive experiments show that it improves the SOTA performance on few-shot image classification by considerable margins (2-5\%). Furthermore, the GML solution is more robust than the existing deep learning solutions, in that its performance can consistently improve as the size of query set increases while the performance of its deep learning alternatives remains essentially flat or even becomes worse.
  
  \end{itemize}
  
	
	The rest of this paper is organized as follows: Section~\ref{sec:related work} reviews related work. Section~\ref{sec:Task Statement} defines the task. Section~\ref{sec:The GML Framework} presents the GML framework for few-shot image classification. Section~\ref{sec:Factor Graph Construction} details factor graph construction. Section~\ref{sec:experiments} presents our empirical evaluation results. Finally, Section~\ref{sec:Conclusion} concludes the paper with some thoughts on future work.
	

%% file: 2.related_work.tex
\section{Related Work}\label{sec:related work}



In this section, we review related work from the orthogonal perspectives of few-shot image classification and gradual machine learning.


\subsection{Few-shot Image Classification}

The existing work for few-shot image classification can be broadly categorized into two groups: inductive learning and transductive learning.

\textbf{Inductive Learning.}
The approach of inductive learning typically trains a neural model by labeled training samples, and then uses the learnt model to separately classifies each unlabeled sample in test classes. It focuses on learning high-quality and transferable features with a CNN backbone. The most commonly used backbones include ResNet-12~\cite{bendou2022easy,zhang2020deepemd,afrasiyabi2022matching,wertheimer2021few}, ResNet-18~\cite{2016Matching,2017Prototypical,sung2018learning,chen2019closer}, and WRN-28-10~\cite{zhu2022ease,hu2022squeezing,hu2021leveraging}. Based on these backbone models, some researchers proposed to further tune the mechanism of feature extraction for improved classification accuracy. For instance, Afeasiyabi et al. proposed to enrich image representations by embedding self-attention mappers in different convolutional layers~\cite{afrasiyabi2022matching}. Zhao et al. designed a structure of Self-Guided Information Convolution (SGI-Conv) to guide the extraction of discriminative features~\cite{zhao2022self}. 

Few-shot classifier training is usually conducted by metric-based learning~\cite{jiang2020multi}, which aims to learn a similarity classifier over a feature space. First, it randomly samples labeled training samples and divides them into support and query sets, thereby constructing multiple different episodes. Then, it uses mini-batches of episodes to train an end-to-end network, assuming that the resulting features will be representative of novel test classes. For instance, Vinyals et al. proposed a matching network to learn embedding~\cite{2016Matching}, whereas Snell et al. proposed a prototypical network to build a pre-class prototype representation~\cite{2017Prototypical}. Sung et al. also presented a relation network, which can learn a non-linear distance metric via a simple neural network instead of using a fixed linear distance metric~\cite{sung2018learning}. 

Transfer learning has also been extensively leveraged for few-shot classifier training~\cite{pan2009survey}. It typically follows a standard two-phase process consisting of pre-training and fine-tuning. In the pre-training phase, it learns transferable knowledge or experience, usually in the form of feature extractors, on base classes that contain ample labeled instances. In the subsequent fine-tuning phase, it usually freezes the learned feature extractors, and trains a new classifier on the support set to recognize novel classes~\cite{verma2019manifold,mangla2020charting}. Many solutions for few-shot image classification have incorporated the concept of transfer learning in their architecture design, including baseline++\cite{chen2019closer}, SimpleShot\cite{wang2019simpleshot}, RFS\cite{tian2020rethinking} and S2M2\cite{mangla2020charting}.

\textbf{Transductive Learning.}
 Transductive learning usually shares the process of backbone training with inductive learning. However, in the test phase, instead of classifying each unlabeled sample separately, it labels unlabeled samples in a collective manner, aiming to exploit sample relationships for improved performance~\cite{hu2021leveraging,zhu2022ease,bendou2022easy,hu2022squeezing}. 


	
	
	
	To facilitate class information propagation, transductive learning typically leverages both support sets and query sets to construct matrics for sample relationship representation. For instance, Kim et al. proposed to construct a graph, where nodes represented samples and edges represented relationships between samples~\cite{kim2019edge}. Yang et al. instead introduced a dual-graph structure for sample relationship representation~\cite{yang2020dpgn}, in which dual graphs were separately constructed based on two similarity relations. To improve intra-class coherence and uniqueness, the approach of TRPN regarded each support-query pair relationship as a graph node, named the relationship node, and leveraged the known relationships between support samples to guide relationship propagation in the graph~\cite{ma2020transductive}.  Various graph neural networks have also been used for label propagation. For instance, Zhu et al. presented a graph construction network to predict the task-specific graph for label propagation~\cite{zhu2023transductive}. Liu et al. argued that there exists a bias between the prototype representations and the expected representations, and provided a simple strategy to rectify the bias based on intra-class and inter-class assumptions~\cite{liu2020prototype}.

  More recently, the Sinkhorn-based approach has achieved promising results on image classification. Its main idea was to  model class information propagation as a transportation optimization problem~\cite{chobola2021transfer,hu2021leveraging}. Huang et al. first proposed the Sinkhorn K-means~\cite{huang2019few}, which used a prototype for query image classification in an unsupervised manner, whereas zhu at al. extended it to the semi-supervised setting~\cite{zhu2022ease}. To balance data distribution between classes, Hu et al. preprocessed feature vectors such that they conform to a Gaussian distribution. In the following work, they successively presented the MAP~\cite{hu2021leveraging} and Boosted Min-Size Sinkhorn algorithm~\cite{hu2022squeezing} to improve propagation efficiency. Similarly, Shalam and bendou et al. proposed the metric of Earth Mover's distance to capture high-level relationships among data points within classes, further optimizing Sinkhorn-based information propagation~\cite{shalam2022self,bendou2022easy}.


  \subsection{Gradual Machine Learning}

  The non-i.i.d learning paradigm of Gradual Machine Learning (GML) was originally proposed for the task of entity resolution~\cite{hou2019gradual,HouTKDE,zhong2021attention}. It can gradually label instances in the order of increasing hardness without the requirement for manual labeling effort. Since then, GML has been also applied to the task of sentiment analysis~\cite{wang2021aspect,ahmed2021dnn,wang2023supervised}. In the unsupervised setting, even though GML can achieve competitive performance compared with many supervised approaches, its performance is still limited by inaccurate and insufficient knowledge conveyance~\cite{hou2019gradual,HouTKDE,ahmed2021dnn}. In the supervised setting, the efficacy of GML depends on supervised extraction of features by DNNs; its performance may deteriorate significantly if the provided labeled samples are insufficient. Our work in this paper is different from previous GML work in two aspects: 1) it targets the new task of image classification; 2) it investigates the efficacy of GML in the few-shot setting, where only a few labeled samples are available. It can be observed that our proposed solution leverages the existing deep backbones for feature extraction, but uses gradual learning as the new inference engine. Therefore, it can be generally considered as an approach of transductive learning.    

%% file: 3.Task_Statement.tex
\section{Task Statement}\label{sec:Task Statement}


   The task of few-shot image classification is usually defined in the $N$-way $K$-shot manner, in which $N$ and $K$ denote the number of classes and the number of labeled samples per class respectively. A workload consists of three disjoint sets: a training set, $\mathcal{D}_{R}$, a validation set $\mathcal{D}_{V}$ and a test set $\mathcal{D}_{T}$, in which the classes of $\mathcal{D}_R$, $\mathcal{D}_V$ and $\mathcal{D}_T$ are distinct from each other.
The training set of $\mathcal{D}_R$ contains a large number of labeled samples, and is usually used to train a generic feature extractor.
The validation set of $\mathcal{D}_V$, used for model selection, also contains labeled samples. The test set of $\mathcal{D}_T$, used for final evaluation, consists of support sets and query sets. Following the standard few-shot setting, final evaluation is supposed to be performed on a set of $N$-way $K$-shot tasks. Concretely, each task of $T_i$ is composed of a support and a query set. The support set, $\mathcal{S}$ = $\{(x_{k,n}^S, y_{k,n}^S)|k\in [1,K], n\in [1,N]\}$, contains $K$ samples per class, and the query set, $\mathcal{Q}$ = $\{(x_{k,n}^Q, y_{k,n}^Q)|k\in [1,K], n\in [1,N]\}$, contains $Q$ samples per class. Given a test task, a classifier needs to select one class from the provided $N$ candidates for each sample in the query set.

  In this paper, we consider the setting of transductive learning, in which a classifier has access to all the samples in $D_T$ when reasoning about the labels of samples in its query sets.

%% file: 4.The_GML_Framework.tex
\section{The GML Framework}\label{sec:The GML Framework}

As shown in Figure~\ref{fig:model structure}, consistent with the general paradigm, the GML framework for few-shot image classification consists of the following three components:

\subsection{Evidential Instance Labeling}


       GML begins with some initial evidential observations. In the unsupervised setting, initial instance labeling can be achieved by human-crafted rules or unsupervised clustering. For instance, an instance very close to a class centroid usually has a high chance of belonging to the class. Few-shot image classification supposes that the support sets contain some labeled sample images (e.g., 5 images per class for 5-shot learning). Therefore, these labeled images can naturally serve as initial evidential samples, even though their number is very limited. In the case of 5-shot learning, the labeled samples in support sets are considered as initial evidential observations. In the case of 1-shot learning, we expand the set of initial evidential observations by automatically labeling one additional sample per class. Specifically, for each class, GML automatically labels the unlabeled sample image closest to the class centroid. As a result, for 1-shot learning, GML actually begins with 2 evidential observations per class.     
				
\subsection{Feature Extraction and Influence Modeling}
    
		In GML, features serve as the medium for gradual learning. This step extracts the common features shared by the labeled and unlabeled samples. To facilitate extensive knowledge conveyance, it is desirable that a wide variety of features are extracted to capture diverse information. For each extracted feature, this step also needs to model its influence over the class labels of relevant samples.
		
		For image classification, CNN has been shown to be much more effective at extracting discriminative features than previous alternatives (e.g., manually crafted mechanisms). Therefore, our solution leverages CNN backbones to extract implicit image features that are indicative of class status. Specifically, it transforms images into high-dimensional vector representations in an embedding space by a CNN, and then uses the monotonic metrics, e.g., class centroid distance and k-nearest neighborhood, to extract discriminative features. It can be observed that the closer to a class centroid an image is, the more likely it belongs to the class; similarly, the more same-class nearest neighbors an image has, the more likely it belongs to the same class. 
		
\begin{figure}[ht]
\centering
\includegraphics[width=0.5\linewidth]{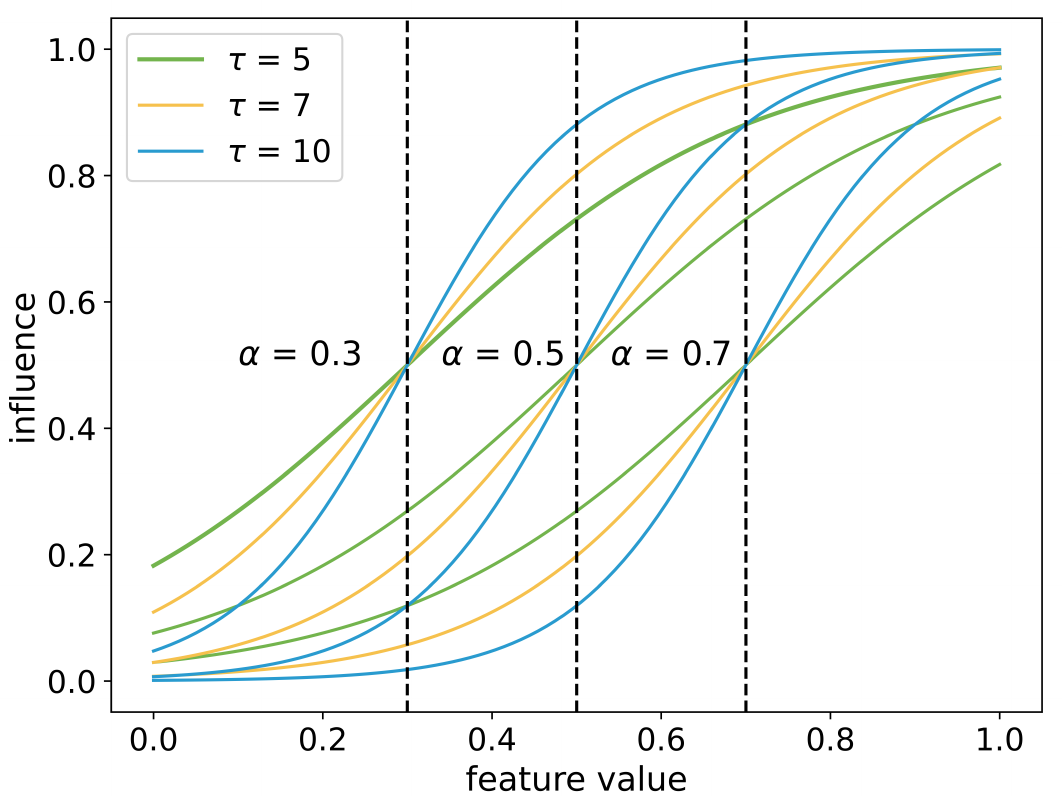}
\caption{The illustrative examples of Sigmoid influence modeling.}
\label{fig:sigmoid}
\end{figure}

		As in previous work, we model a feature's influence over images' class status by a monotonous Sigmoid function. As shown in Figure~\ref{fig:sigmoid}, a sigmoid function has two parameters, $\alpha$ and $\tau$, which denote the midpoint and steepness of the curve respectively. Formally, given a class label, $c$, and its feature, $f_c$, the influence of $f_c$ over the class status of an image, $d$, is represented by
\begin{equation}
\label{eq:sigmoid}
  P_{f_c}(d) = \frac{1}{{1 + {e^{ - {\tau _{f_c}}(x_{f_c}(d) - {\alpha _{f_c}})}}}},
\end{equation}

in which $P_{f_c}(d)$ denotes the probability of $d$ having the label of $c$ as indicated by $f_c$, $x_{f_c}(d)$ represents $d$'s feature value w.r.t $f_c$. According to Eq.~\ref{eq:sigmoid}, provided with the values of $\alpha_{f_c}$ and $\tau_{f_c}$, the influence model statistically dictates that any feature value of $x_{f_c}(d)$ corresponds to a label probability. As shown in Figure~\ref{fig:sigmoid}, different combinations of $\alpha_{f_c}$ and $\tau_{f_c}$ can result in different influence model shapes. Typically, the value of $P_{f_c}(d)$ increases with the feature value of $x_{f_c}(d)$.


		  

\subsection{Gradual Inference}

		GML fulfills gradual learning by iterative factor inference on a factor graph, $G$, which consists of a set of evidence variables, a set of inference variables, and a set of factors. For few-shot image classification, an evidence variable represents a labeled image in the support set, an inference variable represents an unlabeled image in the query set, and a factor represents the correlation between images. Typically, GML labels only one image at each iteration. Once an inference variable is labeled, its value remains unchanged and would serve as an evidence variable in the following iterations. 

    Formally, we denote the set of evidence variables by ${\bf{\Lambda}}$, the set of inference variables by $\bf{V_I}$, and the group of factor functions of variables indicating their correlations by ${\phi_{w_i}}(V_i)$. In the case of few-shot image classification, each variable in the factor graph is supposed to take one of several distinct values, each of which corresponds to a class label. 
Then, the joint probability distribution over $V=\{\Lambda, V_I\}$ of $G$ can be formulated as 
	\begin{equation}
		\label{eq:joint_prob}
		\begin{split}
			P_{\bf{w}}(\Lambda, V_I)=&\frac{1}{Z_{\bf{w}}}\prod_{i=1}^m\phi_{w_i}(V_i)
		\end{split}
	\end{equation}
	where $V_i$ denotes a set of variables, $w_i$ denotes a factor weight, $m$ denotes the total number of factors and $Z_{\bf{w}}$ denotes the normalization constant. Factor inference on $G$ learns factor weights by minimizing the negative log marginal likelihood of evidence variables as follows:
	\begin{equation} \label{eq:weight-learning}
		\hat {\bf{w}}  = arg \min \limits_{\bf{w}} -log \sum_{V_I} P_{\bf{w}}(\Lambda, V_I).
	\end{equation}

 	In each iteration, GML typically labels the inference variable with the highest degree of evidential certainty, which is measured by the inverse of entropy as follows
	 \begin{equation}
 E(v)=\frac{1}{H(v)}, 
 \end{equation}
where 
\begin{equation}
	H(v) = - (P_{max}({v})  \cdot {\log _2}P_{max}({v}) + 
	    (1-P_{max}({v}))  \cdot (1-{\log _2}P_{max}({v}))), 
	\end{equation}
where $E(v)$ and $H(v)$ denote the evidential certainty and entropy of $v$ respectively, and $P_{max}(v)$ denotes the max estimated class probability of $v$. 


\begin{algorithm}[t]
	\caption{Scalable Gradual Inference.}
	\label{alg:gradualinference}
	\While{there exists any unlabeled variable in $G$}
	{
		$V' \leftarrow$ all the unlabeled variables in $G$\;
		\For{$v\in V'$}
		{
			Measure the evidential support of $v$ in $G$\;
		}
		Select top-$m$ unlabeled variables with the most evidential support (denoted by $V_m$) \;
		\For{$v\in V_m$}
		{
			Approximately rank the entropy of $v$ in $V_m$\;
		}
		Select top-$n$ most promising variables in terms of entropy in $V_m$ (denoted by $V_n$) \;
		\For{$v\in V_n$}
		{
			Compute the probability of $v$ in $G$ by factor graph inference over a subgraph of $G$\;
		}
		Label the variable with the minimal entropy in $V_n$\;
	}
\end{algorithm}

To improve efficiency, as usual, the GML solution implements gradual inference by a scalable approach as sketched in Algorithm~\ref{alg:gradualinference}, which is essentially the same as what was previously proposed for entity resolution and sentiment analysis~\cite{hou2018r,wang2021aspect}. Scalable gradual inference consists of three steps: 1) measurement of evidential support; 2) approximate ranking of entropy; 3) subgraph factor inference. In the first step, it selects the top-$m$ unlabeled variables with the most evidential support in $G$ as the inference candidates. For each unlabeled variable, GML measures its evidential support from each feature by the degree of labeling confidence indicated by labeled observations and then aggregates them based on the Dempster-Shafer theory\footnote{https://en.wikipedia.org/wiki/Dempster-Shafer\_theory}. In the second step, it approximates entropy estimation by an efficient algorithm on the $m$ candidates and selects only the top-$n$ most promising variables among them for factor graph inference. Finally, the third step estimates the class probabilities of these $n$ selected variables by factor graph inference. 
	
It is noteworthy that the open-sourced GML engine has standardized the process of scalable gradual inference\footnote{https://chenbenben.org/gml.html}. Therefore, our GML implementation of few-shot image classification only needs to construct a factor graph, while scalable gradual inference on the factor graph can be automatically executed by the engine. Therefore, in the following section, we focus on how to construct the factor graph for few-shot image classification.

%% file: 5.Factor_Graph_Construction.tex
\section{Factor Graph Construction}\label{sec:Factor Graph Construction}

In this section, we first describe how to extract discriminative features by deep neural networks, and then elaborate how to model them as factors in a factor graph to enable gradual learning.

\subsection{Discriminative Feature Extraction}\label{sec:Ifve}


Our solution essentially trains a CNN model to learn high-dimensional vector representations of images in a class-sensitive embedding space, and then extracts monotonic features based on the learned representations to capture their class correlation. Since CNNs are prone to overfitting due to the limited amount of labeled images, we use two different backbones, ResNet-12 and WRN-28-10, for discriminative feature extraction to ensure feature diversity.  
	
	
	
\begin{figure*}[htbp]
    \centering
    \includegraphics[width=0.9\textwidth]{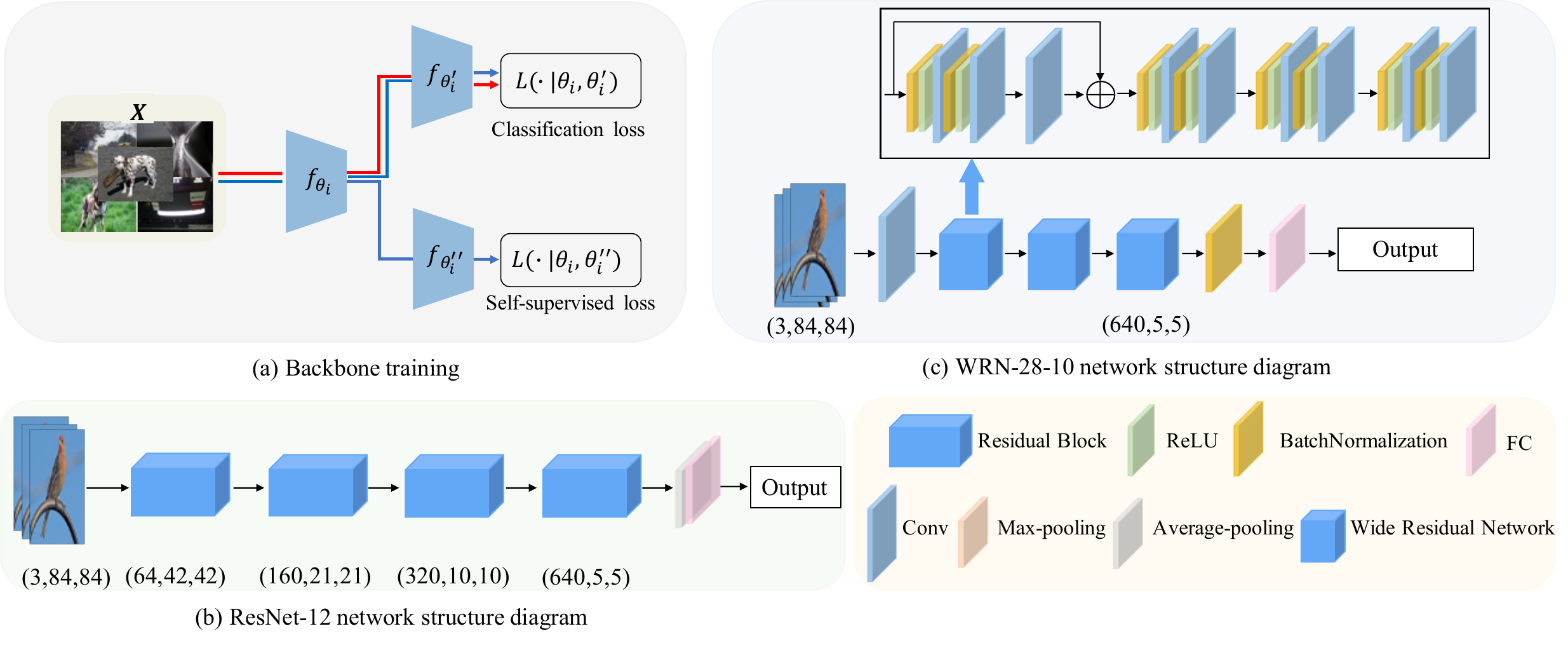}
    \caption{ Backbone training: (a) the Y-shaped training model; (b) the network structure of ResNet-12; (c) the network structure of WRN-28-10.}
		\label{fig:network structure}
\end{figure*}

We have sketched the network structures of ResNet-12 and WRN-28-10 in Figure~\ref{fig:network structure} (b) and (c) respectively. ResNet-12 consists of four residual blocks, each of which contains three convolutional layers. Based on ResNet, the WRN-28-10 (Wide Residual Network) architecture increases the width of residual blocks from 12 to 28. By increasing the width of residual blocks, WRN-28-10 attains more advanced representational capabilities, and thus can capture finer-granular image features. In ResNet-12, we extract vector representations by the output of Average pooling2D and the input of the FC layer. In WRN-28-10, we extract vector representations by the output of the BatchNormalization layer and the input of the FC layer by the backbones. Both extracted vectors have the dimension of 640. Our ablation study has shown that using both ResNet-12 and WRN-28-10 for feature extraction is considerably better that using either of them. 

We train our backbones using the methodology called EASY presented in~\cite{bendou2022easy}. Its basic principle is to take a standard classification architecture, and then set up two classifiers after the second-to-last layer of the network: one classifier for identifying the class of input samples and a new logistic regression classifier which can determine which one of the four possible rotations (one-quarter rotations of 360°) was applied to the input samples. EASY uses a two-step process to train model. In the first step, samples are directly input to the model and its first classifier. In the second step, samples are arbitrarily rotated and separately input into the two classifiers. Once training is complete, we freeze the backbone and use it, $f(\theta)$ as shown in Figure~\ref{fig:network structure} (a), to extract representation vectors for the images in $\mathcal{D}_V$ and $\mathcal{D}_T$.


Then, based on the learned vector representations, we extract two types of monotonic features as follows:
	
\begin{itemize}
  \item {\bf Class Centroid Distance (\emph{CCD}).} We estimate the prototype class centroid of each class by its support set, and then measure an image's similarity with a class by calculating its distance to the class centroid, which is defined as 1.0 minus vector cosine similarity. It is obvious that the smaller the distance is, the more likely the image belongs to the class. We denote the unary feature of class centroid distance by \emph{CCD}. 
	
	
	\item {\bf K-nearest Neighborhood (\emph{KNN}).} Since a CNN classifier tends to separate the images with different class labels as far as possible while clustering the images with the same label, two images appearing very close in its corresponding embedding space usually have the same label. Therefore, we extract k-nearest neighborhood relations, ($v_i$, $v_j$, $sim_{i,i}$), in which $sim_{i,j}$ denotes the cosine similarity between the vector representations of $v_i$ and $v_j$. We denote the binary feature of k-nearest neighborhood by \emph{KNN}. In practical implementation, we suggest to set the value of $k$ within the reasonable range of [5,7].
\end{itemize}	

\begin{figure*}[htbp]
  \centering
  \includegraphics[width=0.8\textwidth]{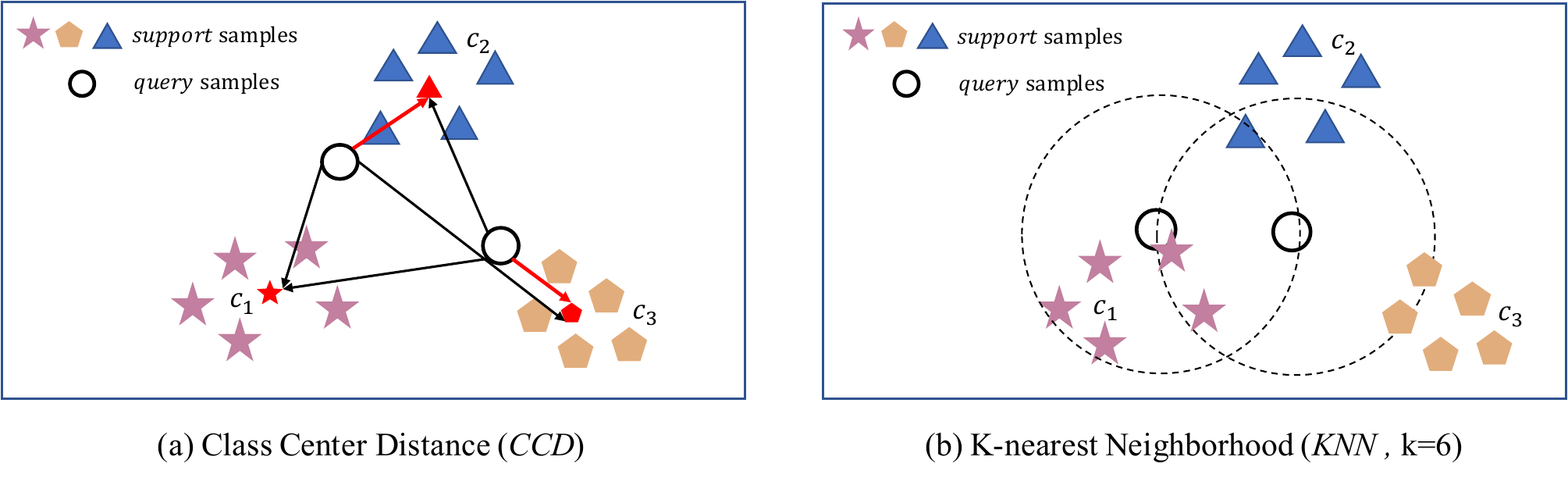}
  \caption{The visualization of \emph{CCD} and \emph{KNN} features in the case of 5-way 5-shot, in which red dots represent class centroids and black circles represent unlabeled samples: (a) the unary feature of Class Centroid Distance (\emph{CCD}) is constructed by measuring an unlabeled sample's distance to a class centroid; (b) the binary feature of K-nearest Neighborhood (\emph{KNN}) is constructed by measuring the cosine similarity between an unlabeled sample and its k-nearest neighbors in the embedding space.}\label{fig:distance structure}
\end{figure*}

We have visualized the unary \emph{CCD} and the binary \emph{KNN} features in Figure~\ref{fig:distance structure} (b) and (c) respectively. 
It is noteworthy that \emph{CCD} and \emph{KNN} are complementary to each other, in that \emph{CCD} captures an image's correlation with global representatives, the prototype class centroids, while \emph{KNN} captures an image's correlation with local representatives, its nearest neighbors. 


\subsection{Feature Influence Modeling}\label{sec:GML}

 Since both \emph{CCD} and \emph{KNN} features are monotonic w.r.t class probability, we model their influence over class status by the sigmoid function as shown in Figure~\ref{fig:sigmoid}. Formally, denoting a \emph{CCD} feature by $f_c$, in which $c$ denotes a class, we model the influence of $f_c$ over a variable, $v$, by a unary factor defined as follows: 
\begin{equation}
	\label{eq:unary factor}
	\varphi_{f_c}(v) = 
	\left \{
	\begin{array}{ll}
	e^{w_{f_c}(v)}      &    
     if $ $ v = c;  \\
  1 &  if $ $ v\neq  c.
	\end{array} 
\right.
\end{equation}
where $w_{f_c}(v)$ denotes the factor weight of $v$, and 

\begin{equation}
\label{eq:unaryfactorweight}
  w_{f_c}(v) = \theta_{f_c}(v)\cdot \tau_{f_c}\cdot (x_{f_c}(v) - {\alpha_{f_c})},
\end{equation}
in which $\theta_{f_c}(v)$ denotes the confidence on influence modeling of $f_c$, $x_{f_c}(v)$ denotes the feature value of $v$, or the distance to the class centroid of $c$, and $\tau_{f_c}$ and $\alpha_{f_c}$ denote the steepness and mid-point of a sigmoid function respectively. In our implementation, as in previous GML work~\cite{hou2019gradual,HouTKDE}, we estimate $\theta_{f_c}(v)$ by the theory of regression error bound. The parameter values of $\tau_{f_c}$ and $\alpha_{f_c}$ are however supposed to be continuously optimized based on evidential observations in the process of gradual learning.

	Similarly, we model the influence of the \emph{KNN} feature by the following binary factor: 
\begin{equation}
	\label{eq:binary factor}
	\varphi_{f_b}(v_i, v_j) = 
	\left \{
	\begin{array}{ll}
		e^{w_{f_b}(v_i,v_j)}     &      if \ v_i = v_j; \\
		1    &   otherwise.
	\end{array} 
	\right.
\end{equation}	
in which $f_b$ denotes the binary \emph{KNN} feature, $v_i$ and $v_j$ denote the two variables sharing the feature of $f_b$, $w_{f_b}(v_i,v_j)$ denotes the factor weight, and
\begin{equation}
\label{eq:unaryfactorweight}
  w_{f_b}(v_i,v_j) = \theta_{f_u}(v_i,v_j)\cdot \tau_{f_b}\cdot (x_{f_b}(v_i,v_j) - {\alpha_{f_b}}), 
\end{equation}
in which $\theta_{f_u}(v_i,v_j)$ denotes the confidence on binary feature influence modeling, $x_{f_b}(v_i,v_j)$ denotes the vector similarity of $v_i$ and $v_j$, and $\tau_{f_b}$ and $\alpha_{f_b}$ denote the steepness and mid-point of a sigmoid function. Similar to the unary factor, we estimate $\theta_{f_u}(v_i,v_j)$ by the theory of regression error bound, while $\tau_{f_b}$ and $\alpha_{f_b}$ need to be 
continuously optimized in the process of gradual learning. 

%% file: 6.Experiments.tex
\section{Empirical Evaluation}\label{sec:experiments}

In this section, we empirically evaluate the performance of our proposed solution by a comparative study on real benchmark datasets. It is organized as follows: Subsection~\ref{sec:setup} describes the experimental setup. Subsection~\ref{sec:comparison} presents the comparative evaluation results. Subsection~\ref{sec:ablation} presents the evaluation results of ablation study. Finally, Subsection~\ref{sec:sensitivity} evaluates the performance sensitivity of GML w.r.t key parameters.

\subsection{Experimental Setup} \label{sec:setup}

  We use four widely used benchmark datasets in our empirical study: 
\begin{itemize}
\item \textbf{MiniImageNet\cite{ravi2017optimization}}: it contains totally 100 classes, each of which contains 600 images with a size of 84$\times$84. The classes are split among training, validation, and test sets by the ratio of (64:16:20);

\item \textbf{TieredImageNet\cite{ren2018meta}}:it was created by selecting 34 categories from the ILSVRC-2012 imagenet, with each superclass containing 10-30 subclasses. There are totally 20 superclasses (351 subclasses) on the training set, 6 superclasses (97 subclasses) in the validation set and 8 superclasses (160 subclasses) in the test set. All the images are of the size of 84$\times$84; 

\item \textbf{Cifar-FS\cite{bertinetto2018meta}}: it contains totally 100 classes, each of which contains 600 images with the size of 32$\times$32. The classes are split among training, validation and test sets by the ratio of (64:16:20);

\item \textbf{CUB-200-2011\cite{wah2011caltech}}: also known as Caltech-UCSD Birds-200-2011, it contains totally 200 bird species, which are split among training, validation, and test sets by the ratio of (100:50:50).
\end{itemize}	


 We compare the proposed GML solution with the SOTA methods of both inductive learning and transductive learning: 1) the methods of inductive learning include recently proposed ProtoNet~\cite{2017Prototypical}, MetaQDA~\cite{zhang2021shallow}, SetFeat-12~\cite{afrasiyabi2022matching} and PFENet~\cite{zhao2022self}. Among them, PFENet reported the overall best performance; 2) the methods of transductive learning include recently proposed TPN~\cite{sung2018learning}, LaplacianShot~\cite{ziko2020laplacian}, COM-FSC~\cite{liu2023cycle}, DFMN-MCT~\cite{kye2020meta}, PT-MAP~\cite{hu2021leveraging}, EASE+SIAMESE~\cite{zhu2022ease}, EASY~\cite{bendou2022easy} and PEM$_n$E-BMS*~\cite{hu2022squeezing}. Among them, the most recently proposed models, e.g., EASY, EASE+SIAMESE and PEM$_n$E-BMS*, have reported highly competitive performance. 

  Due to the large number of the compared methods, we directly compare the results of GML in term of accuracy with the results that have been reported in these methods' original papers. For fair comparison, on each workload, as usual, we report the average and standard variance over 10000 rounds of testing. Since most of the existing methods only report results on 2-3 test datasets of the 4 datasets we have used, we mark a compared method's result on a dataset as null (-) if it was not reported in the original paper.       

  For comparative evaluation, we have compared performance in both scenarios of intro-domain classification, where training and test sets come from the same data source, and cross-domain classification, where training and test sets come from different sources, e.g., a model is trained on MiniImageNet but tested on CUB-200-2011. Obviously, cross-domain classification is more challenging than intro-domain classification. In practical scenarios, it is usually expensive to manually label samples, but much easier to retrieve unlabeled samples. Therefore, we have also evaluated the robustness of the proposed GML solution by increasing the size of query set, or the samples to be labeled in the query set, and compared its performance with the existing SOTA alternatives.

We have implemented the proposed solution based on the open-sourced GML engine\footnote{https://github.com/gml-explore/gml}. In the GML implementation, we leverage both ResNet-12 and WRN-28-10 to extract deep vector representations. In the generation of \emph{KNN} features, by default, we set $k=6$. In the implementation of gradual inference, by default, we set the number of candidates with the highest evidential support at 50, and the number of candidates with the smallest approximate entropy at 10.  At each iteration of gradual inference, GML labels the 10 samples with the smallest approximate entropy by factor inference. After each iteration, given the newly labeled images, the algorithm correspondingly updates evidential support and approximate entropy estimation. In our sensitivity evaluation, we will show that the performance of GML is very stable w.r.t these parameters provided that their values are set within reasonable ranges. We have open-sourced the GML implementation\footnote{https://github.com/chn05/FSIC\_GML}.



\subsection{Comparative Evaluation} \label{sec:comparison}


\textbf{Intro-domain classification:}
the comparative evaluation results have been presented in Table~\ref{table:minitiered} and~\ref{table:cifar-fs_cub}. It can be observed that the transductive approaches consistently perform considerably better than their inductive alternatives on all the workloads. It is worth pointing out that GML consistently outperforms the best transductive alternative by considerable margins on all the workloads. For instance, on MiniImageNet and TierImageNet, in the case of 5-way 1-shot learning, GML beats EASY+SIMESE, which is the best transductive approach based on the reported results, by the margins of 2.79\% and 2.41\% respectively. In the case of 5-way 5-shot learning, the improvement margins are 4.65\% and 1.5\% respectively. On Cifar-FS and CUB-100-2011, PEM$_n$E-BMS* is instead the best transductive approach. In the case of 5-way 1-shot learning, GML outperforms PEM$_n$E-BMS* by the margins of 4.58\% and 3.72\% respectively. In the case of 5-way 5-shot learning, the improvement margins are 3.89\% and 2.93\% respectively. 

\begin{table*}[htbp]
    \centering
    \caption{Comparative results on the MiniImageNet and TieredImageNet datasets: GML clearly achieves the SOTA performance on both datasets, and the margins are considerable.}\label{table:minitiered}
    \small
    \begin{tabular}{c|c|c|c|c|c}
    \toprule
    \multicolumn{2}{c|}{\textbf{DataSets}} & \multicolumn{2}{c|}{\cellcolor[HTML]{9698ED}{\textbf{Mini-ImageNet}}} & \multicolumn{2}{c}{\cellcolor[HTML]{FFCE93}{\textbf{Tiered-ImageNet}}} \\ 
    \midrule
    Setting & Methods & 5-way 1-shot(\%) &5-way 5-shot(\%) & 5-way 1-shot(\%) & 5-way 5-shot(\%)\\ 
    \midrule
    \multirow{10}{*}{\textbf{Inductive}} &  Relation\cite{sung2018learning}  &$52.48\pm0.86$ & $69.83\pm0.68$  & $-$ & $-$  \\
    & Baseline++\cite{chen2019closer}  &  $53.97\pm0.79$ & $75.90\pm0.61$  & $-$ & $-$\\
    &MatchingNet\cite{2016Matching} &   $52.91\pm0.88$ & $68.88\pm0.69$ & $-$ & $-$ \\
    &ProtoNet\cite{2017Prototypical} &  $54.16\pm0.82$ & $73.68\pm0.65$ &  $65.65\pm0.92$ & $83.40\pm0.65$ \\
    &$S2M2_R$~\cite{mangla2020charting} &  $64.93\pm0.18$ & $83.18\pm0.11$ & $73.71\pm0.22$ & $88.59\pm0.14$\\
    &DeepEMD\cite{zhang2020deepemd} &  $65.91\pm0.82$ & $82.41\pm0.56$ & $71.16\pm0.87$ & $86.03\pm0.58$ \\
    &FRN\cite{wertheimer2021few} &   $66.45\pm0.19$ & $82.83\pm0.13$ & $71.16\pm0.22$ & $86.01\pm0.15$\\
    &MetaQDA\cite{zhang2021shallow} &  $67.83\pm0.64$ & $84.28\pm0.69$ & $74.33\pm0.65$ & $89.56\pm0.79$\\
    &SetFeat12\cite{afrasiyabi2022matching} &  $68.32\pm0.62$ & $82.71\pm0.46$ & $73.63\pm0.88$ & $87.59\pm0.57$ \\
    &PFENet\cite{zhao2022self} &  $68.76\pm0.75$ & $84.67\pm0.52$ & $74.93\pm0.84$ & $89.62\pm0.50$ \\
    \midrule
    \multirow{10}{*}{\textbf{Transductive}} &TPN\cite{sung2018learning} & $55.51\pm0.86$ & $69.86\pm0.65$ &  $59.91\pm0.94$ & $73.30\pm0.75$\\
    &COM-FSC\cite{liu2023cycle} &   $68.92\pm0.72$ & $85.37\pm0.49$ &  $79.69\pm0.74$ & $90.57\pm0.45$\\
    &LaplacianShot\cite{ziko2020laplacian} & $75.57\pm0.19$ & $84.72\pm0.13$ &  $80.30\pm0.22$ & $87.93\pm0.15$\\
    &DFMN-MCT\cite{kye2020meta} &  $78.55\pm0.86$ & $86.03\pm0.42$ &  $80.89\pm0.84$ & $87.30\pm0.49$\\
    &Transd-CNAPS+FETI\cite{bateni2022enhancing} &   $79.90\pm0.80$ & $91.50\pm0.40$ & $73.80\pm0.10$ & $87.70\pm0.60$ \\
    &PT-MAP\cite{hu2021leveraging} &   $82.92\pm0.26$ & $88.82\pm0.13$ & $85.67\pm0.26$ & $90.45\pm0.14$\\
    &EASE+SIAMESE\cite{zhu2022ease} &$83.00\pm0.21$ & $88.92\pm0.13$ & $88.96\pm0.23$ & $92.63\pm0.13$\\
    &EASY\cite{bendou2022easy} & $84.04\pm0.23$ & $89.14\pm0.11$ &  $84.29\pm0.24$ & $89.76\pm0.14$\\
    &PEM$_n$E-BMS*\cite{hu2022squeezing} &  $83.35\pm0.25$ & $89.53\pm0.13$ & $86.07\pm0.75$ & $91.09\pm0.14$\\
    \rowcolor[HTML]{FFFFFF}&\cellcolor[HTML]{96FFFB} \textbf{GML}  & \cellcolor[HTML]{96FFFB} \textbf{$85.79\pm0.32$} & \cellcolor[HTML]{96FFFB} \textbf{$93.57\pm0.25$} & \cellcolor[HTML]{96FFFB}  \textbf{$91.37\pm0.42$} & \cellcolor[HTML]{96FFFB} \textbf{$94.13\pm0.13$}\\
    \bottomrule
    \end{tabular}
    \end{table*}

\begin{table*}[htbp]
    \centering
    \caption{Comparative results on the Cifar-FS and CUB-100-2011 datasets: GML clearly achieves the SOTA performance on both datasets, and the margins are considerable.}\label{table:cifar-fs_cub}
    \small
    \begin{tabular}{c|c|c|c|c|c}
    \toprule
    \multicolumn{2}{c|}{\textbf{DataSets}} & \multicolumn{2}{c|}{\cellcolor[HTML]{C0C0C0}{\textbf{Cifar-FS}}}  &  \multicolumn{2}{c|}{\cellcolor[HTML]{FE996B}{\textbf{CUB-100-2011}}} \\ 
    \midrule
    \textbf{Setting} &\textbf{Method} & \textbf{5-way 1-shot(\%)} & \textbf{5-way 5-shot(\%)} & \textbf{5-way 1-shot(\%)} & \textbf{5-way 5-shot(\%)} \\ 
    \midrule
    \multirow{9}{*}{\textbf{Inductive}} & MatchingNet\cite{2016Matching} & $43.88\pm0.75$ & $57.05\pm0.76$ & $-$ & $-$\\
    &ProtoNet\cite{2017Prototypical} & $41.54\pm0.76$ & $57.08\pm0.76$ & $-$ & $-$\\
    &DeepEMD\cite{zhang2020deepemd} &$46.47\pm0.78$ & $63.22\pm0.71$ &$-$ & $-$\\
    &SetFeat12\cite{afrasiyabi2022matching} &  $-$ & $-$ &  $79.60\pm0.80$ & $90.48\pm0.44$ \\
    &$S2M2_R$\cite{mangla2020charting} & $74.81\pm0.19$ & $87.47\pm0.13$ & $80.68\pm0.81$ & $90.85\pm0.44$\\
    &RENet\cite{2021Relational} &   $74.51\pm0.46$ & $86.60\pm0.32$ &  $79.49\pm0.44$ & $91.11\pm0.24$\\
    &FRN\cite{wertheimer2021few} & $-$ & $-$ &  $83.55\pm0.19$ & $92.92\pm0.10$ \\
    &PFENet\cite{zhao2022self} &  $-$ & $-$ & $86.09\pm0.19$ & $93.15\pm0.10$ \\
    &MetaQDA\cite{zhang2021shallow} &  $75.83\pm0.88$ & $88.79\pm0.75$ & $-$ & $-$\\
    \midrule
    \multirow{7}{*}{\textbf{Transductive}} & COM-FSC\cite{liu2023cycle} & $-$ & $-$ &  $83.93\pm0.66$ & $93.95\pm0.30$\\
    &EASY\cite{bendou2022easy} & $87.16\pm0.21$ & $90.47\pm0.15$ &  $90.56\pm0.19$ & $93.79\pm0.10$\\
    &iLPC\cite{Lazarou_2021_ICCV}& $86.51\pm0.23$ & $90.60\pm0.48$ &  $91.03\pm0.63$ & $94.11\pm0.30$\\
    &EASE+SIAMESE\cite{zhu2022ease} &  $87.60\pm0.23$ & $90.60\pm0.16$ & $91.68\pm0.19$ & $94.12\pm0.09$\\
    &PT-MAP\cite{hu2021leveraging} &  $87.69\pm0.23$ & $90.68\pm0.15$ & $91.55\pm0.19$ & $93.99\pm0.10$\\
    &PEM$_n$E-BMS*\cite{hu2022squeezing} & $87.83\pm0.22$ & $91.20\pm0.15$ & $91.91\pm0.18$ & $94.62\pm0.09$\\
    \rowcolor[HTML]{FFFFFF}  & \cellcolor[HTML]{96FFFB} \textbf{GML} & \cellcolor[HTML]{96FFFB} \textbf{$92.41\pm0.32$} & \cellcolor[HTML]{96FFFB} \textbf{$95.09\pm0.18$} &  \cellcolor[HTML]{96FFFB} \textbf{$95.63\pm0.06$} & \cellcolor[HTML]{96FFFB} \textbf{$97.55\pm0.43$}\\
    \bottomrule
    \end{tabular}
    \end{table*}
    
	Even with the existing inductive and transductive solutions being considered as a whole, GML consistently improves the reported SOTA performance on the four workloads by considerable margins. In the case of 5-way 1-shot learning, the improvement margins over the SOTA results are 1.75\%, 2.41\%, 4.58\%, and 3.95\% respectively. In the case of 5-way 5-shot learning, the improvement margins are 2.07\%, 1.5\%, 3.89\%, and 2.93\% respectively. Due to the widely recognized challenge of few-shot learning, these margins are truly considerable. Since our GML solution extracts discriminative features by the same deep neural models leveraged by the existing solutions, these evaluation results clearly demonstrate that compared with the existing transductive alternatives, gradual inference is a more effective mechanism for few-shot learning.       


\begin{table}[htbp]
    \centering
    \caption{Comparative results on cross-domain classification: the models are trained on MiniImageNet but tested on CUB-100-2011.}\label{table:cross}
    \small
    \begin{tabular}{c|c|c|c}
    \toprule
    \textbf{DataSets} & \multicolumn{3}{c}{\textbf{MiniImageNet$\rightarrow$CUB-100-2011 (5-way)}} \\
    \toprule
    \textbf{Setting} & \textbf{Methods}  &\textbf{1-shot(\%)} &\textbf{5-shot(\%)} \\ 
    \midrule
    \multirow{4}{*}{\textbf{Inductive}} &PFENet\cite{zhao2022self}   &$48.27$ & $69.51$ \\
    &$S2M2_R$\cite{mangla2020charting} & $48.24$ & $70.44$ \\
    &MetaQDA\cite{zhang2021shallow}  &  $53.75$ & $71.84$\\
    &FRN\cite{wertheimer2021few} & $54.11$ & $77.09$ \\
    \midrule
    \multirow{4}{*}{\textbf{Transductive}} &LaplacianShot\cite{ziko2020laplacian} & $55.46$ & $66.33$ \\
    &COM-FSC\cite{liu2023cycle} &  $53.14$ & $73.02$ \\
    &PEM$_n$E-BMS*\cite{hu2022squeezing}& $63.00$ & $79.15$ \\
    \rowcolor[HTML]{FFFFFF}  & \cellcolor[HTML]{96FFFB} \textbf{GML} &\cellcolor[HTML]{96FFFB} $\textbf{67.29}$ & \cellcolor[HTML]{96FFFB} $\textbf{82.81}$ \\
    \bottomrule
    \end{tabular}
\end{table}

\textbf{Cross-domain classification:} 
 cross-domain classification is usually performed on two datasets containing the same type of objects. Since both MiniImageNet and CUB-100-2011 contain images of bird species, as in previous work~\cite{mangla2020charting,zhang2021shallow,zhao2022self,ziko2020laplacian}, we train models on the MiniImageNet dataset and test its performance on another dataset of CUB-100-2011. 
	
  The comparative evaluation results have been presented in Table~\ref{table:cross}. It can be observed that on both cases of 1-shot and 5-shot learning, GML outperforms the existing alternatives by considerable margins. Specifically, on 1-shot learning, GML beats PEM$_n$E-BMS*, which is the best approach among the existing alternatives, by 4.29\% in terms of accuracy. On 5-shot learning, GML's improvement margin over PEM$_n$E-BMS* is similarly large at 3.66\%. Our experimental results clearly demonstrate that by gradual learning, the features learned in training classes can be better generalized to unseen classes. 


\begin{figure*}[htbp]
    \centering
    \includegraphics[width=0.9\textwidth]{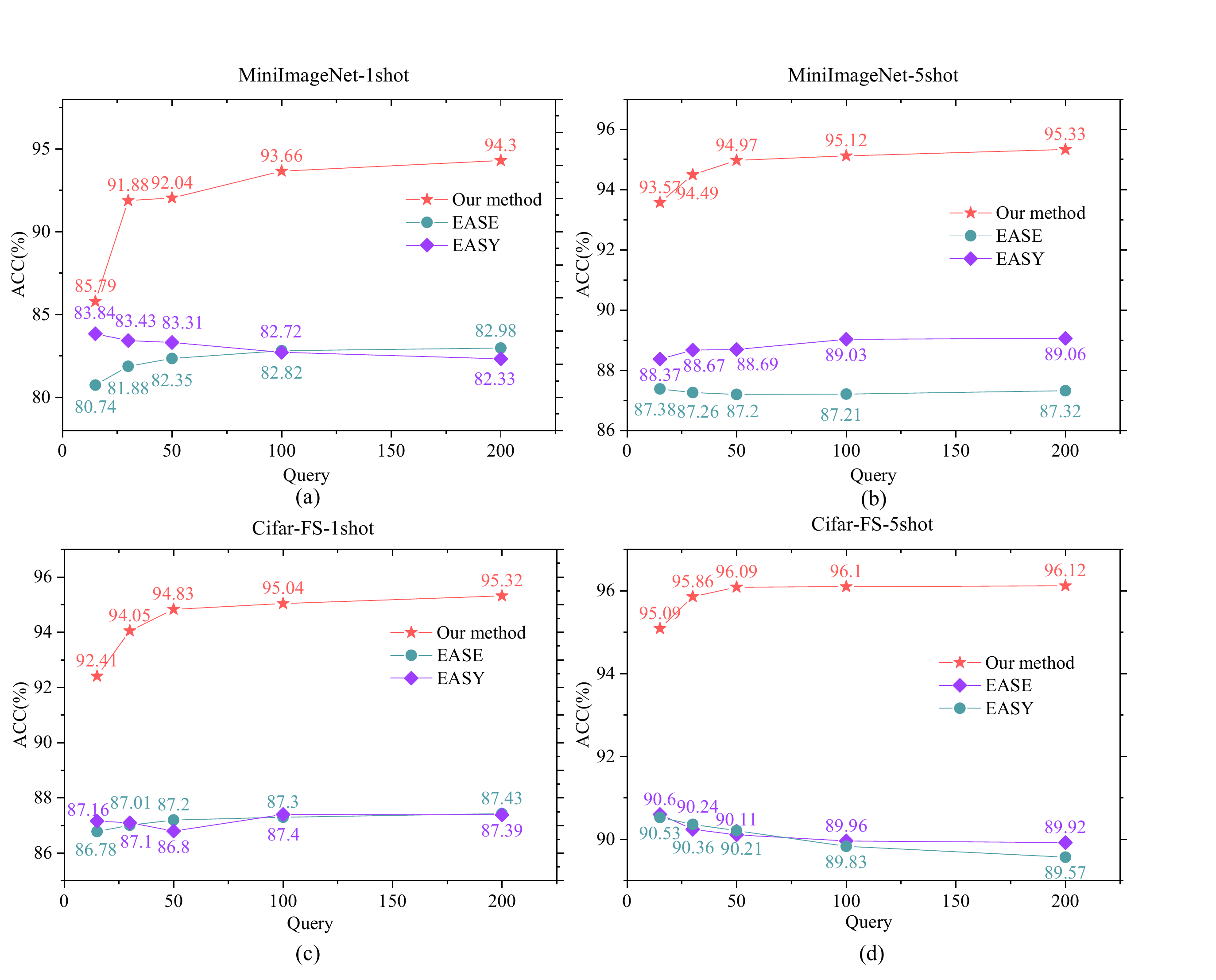}
    \caption{Comparative evaluation results with increasing size of query set on MiniImagenet and Cifar-FS: (a) and (b) represent the accuracy of 1-shot and 5-shot learning on MiniImageNet respectively; (c) and (d) represent the accuracy of 1-shot and 5-shot learning on Cifar-FS respectively.}
		\label{fig:query structure}
\end{figure*}

\begin{table*}[htbp]
    \centering
    \caption{The evaluation result of GML ablation study on MiniImagenet and Cifar-FS: using both ResNet-12 and WRN-28-10 vs using either of them.}\label{table:backbone}
    \small
    \begin{tabular}{c|c|c|c|c|c}
    \toprule
    \multicolumn{2}{c|}{\textbf{DataSets}}  &  \multicolumn{2}{c|}{\textbf{MiniImageNet}} &\multicolumn{2}{c}{\textbf{Cifar-FS}} \\
    \midrule
    \textbf{Methods} & \textbf{Network} & \textbf{5-way 1-shot(\%)} & \textbf{5-way 5-shot(\%)}  & \textbf{5-way 1-shot(\%)} & \textbf{5-way 5-shot(\%)} \\ 
    \midrule
    \multirow{3}{*}{\textbf{GML}} &  ResNet-12 & $84.74$ & $89.50$ & $85.69$ & $89.43$\\
    &  WRN-28-10 & $83.28$ & $88.80$ & $84.06$ & $88.35$\\
    \rowcolor[HTML]{FFFFFF}  &  \cellcolor[HTML]{96FFFB} ResNet-12+WRN-28-10 &\cellcolor[HTML]{96FFFB}$\textbf{85.79}$ &\cellcolor[HTML]{96FFFB}$\textbf{93.57}$ & \cellcolor[HTML]{96FFFB}$\textbf{92.41}$ &\cellcolor[HTML]{96FFFB}$\textbf{95.09}$\\
  
    \bottomrule
    \end{tabular}
    \end{table*}

\begin{table*}[htbp]
        \centering
        \caption{ Parameter sensitivity evaluation results.}\label{table:sensitivity}
        \small
        \begin{tabular}{c|c|c|c|c|c}
        \toprule
        \multicolumn{2}{c|}{\textbf{DataSets}} & \multicolumn{2}{c|}{\textbf{MiniImageNet}} & \multicolumn{2}{c}{\textbf{Cifar-FS}} \\
        \midrule
         & \textbf{top-m/n/k}  &\textbf{5-way 1-shot(\%)} & \textbf{5-way 5-shot(\%)}  &\textbf{5-way 1-shot(\%)} & \textbf{5-way 5-shot(\%)} \\ 
        \midrule
        \rowcolor[HTML]{FFFFFF} \multirow{2}{*}{\textbf{$w.r.t$ m (n=10,k=6)}}  & m=40 &  $85.61$ & $ 93.28$  &  $92.16$ & $95.12$ \\
         & m=60 &  $ 85.56 $ & $ 93.23 $ &  $92.31$ & $ 94.59$  \\
         \midrule
         \multirow{2}{*}{\textbf{$w.r.t$ n (m=50,k=6)}} & n=8 &  $ 85.39$ & $ 93.48$  &  $92.09$ & $94.95$ \\
         & n=12 & $85.69$ & $ 93.56$  &  $92.24$ & $95.10$ \\
         \midrule
        \multirow{2}{*}{\textbf{$w.r.t$ k (m=50,n=10)}}  & k=5 & $ 85.76 $ & $ 93.33 $  &  $ 92.40 $ & $ 94.98 $   \\
        & k=7 & $ 85.15 $ & $ 93.04 $  &  $ 91.83 $ & $ 94.66 $   \\
        \midrule
        \rowcolor[HTML]{96FFFB}\textbf{GML}  & m=50, n=10, k=6 &$ 85.79 $ &$ 93.57 $  & $ 92.41 $ & $ 95.09 $   \\
        \bottomrule
        \end{tabular}
\end{table*}

\textbf{Comparative evaluation with increasing size of query set:}
in the classical setting of few-shot learning, the number of queries per class is set at 15. Therefore, we increase the number of queries from 15 to 30, 50, 100, and finally up to 200, and compare the performance of GML with two recently proposed approaches, EASY and EASE, which can be considered as the SOTA representatives of the existing transductive approaches. 


The comparative evaluation results on MiniImageNet and Cifar-FS have been presented in Figure~\ref{fig:query structure}. The evaluation results on the other two datasets are similar, thus omitted due to space limit. It can be observed that on both 1-shot and 5-shot learning, the performance of GML consistently improves as the number of queries increases, even though by different margins on different workloads. The common pattern is that the performance of GML initially improves considerably as the number of queries increases from 15 to 30, but then gradually flattens out as it continues to increase. For instance, in the case of 1-shot learning on MiniImageNet, the accuracy improves by the margin of around 6\%. from 85.79\% to 91.88\%, when the number of queries increases from 15 to 30, and then continues to improve, even though by smaller margins, up to 94.3\% when the number of queries reaches 200. In comparison, the performance of EASE and EASY fluctuates only marginally when the number of queries increases. For instance, in the case of 1-shot learning on MiniImageNet, the performance of EASY even deteriorates marginally from 83.84\% to 82.98\%; but on 5-shot learning, its performance instead improves slightly, from 88.37\% to 89.06\%. These observations clearly demonstrate that GML is more robust than the existing transductive alternatives. They bode well for its application in real scenarios.

\subsection{Ablation Study} \label{sec:ablation}

To verify the efficacy of leveraging two distinct backbones, i.e., ResNet-12 and WRN-28-10, for gradual learning, we have conducted an ablation study on the GML approach, which compares the solution using both models with the alternatives using either of them. The evaluation results on MiniImagenet and Cifar-FS have been presented in Table~\ref{table:backbone}. The evaluation results on the two other datasets are similar, thus omitted here. It can be observed that the GML using both of them performs considerably better that the alternatives using either of them. These results clearly demonstrate that even though both ResNet-12 and WRN-28-10 have been constructed based on the ResNet network, they are some extent complementary in feature extraction, and integrating them for knowledge conveyance can effectively improve the performance of gradual learning.

\begin{figure*}[htbp]
  \centering
  \includegraphics[width=0.8\textwidth]{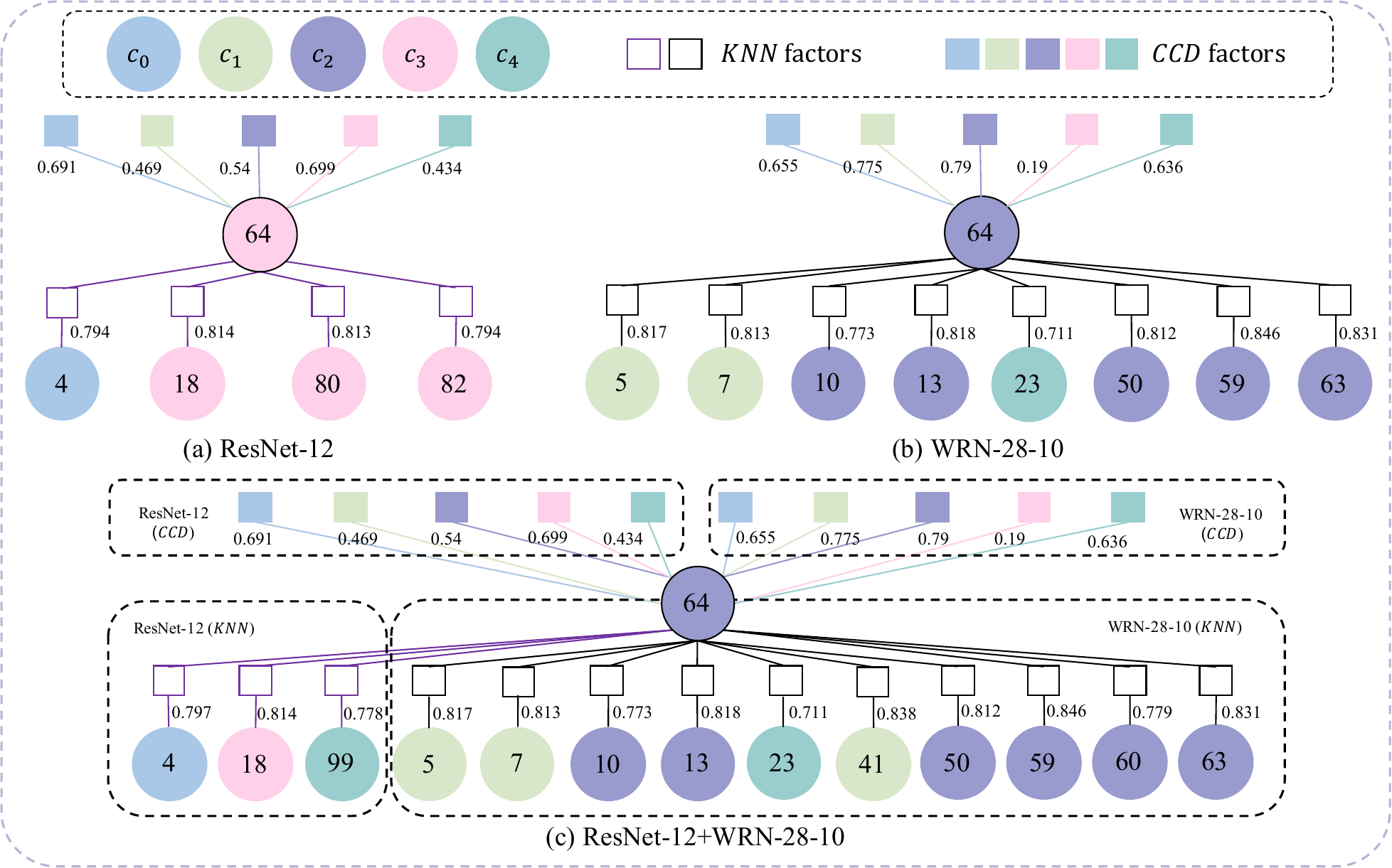}
  \caption{An illustrative example: gradual inference using both ResNet-12 and WRN-28-10 is more effective than using either of them.}
  \label{fig:example}
\end{figure*}

\textbf{An Illustrative Example:} in a run on MiniImageNet, inference accuracy based on ResNet-12 or WRN-28-10 is 85.33\% and 68\% respectively, but the accuracy is better at 94.67\% if gradual inference uses both of them. As shown in Figure~\ref{fig:example}, we take the sample with the id of 64 as an example. In the factor graph constructed based on ResNet, both unary and binary factors point to the class of $c_3$ for the sample. However, based on WRN-28-10, the factors point to its ground-truth class of $c_2$. It can be observed that the fused factor graph constructed based on both ResNet-12 and WRN-28-10 correctly point to the class of $c_2$ while labeling the sample of 64. It is noteworthy that the inference order in different factors may vary. This example clearly demonstrates that the framework of GML can effectively fuse diverse and noisy features to improve gradual knowledge conveyance.

\subsection{Parameter Sensitivity Study} \label{sec:sensitivity}

In this subsection, we evaluate the performance sensitivity of the proposed GML solution w.r.t one key parameter of feature extraction, the $k$ value of k-nearest neighbors (\emph{KNN}) for binary feature extraction, and two key parameters of scalable gradual inference, the number of candidates with the most evidential support and the number of candidates with the smallest approximate entropy, or $m$ and $n$ as shown in Algorithm~\ref{alg:gradualinference}. By default, we set $m$ = 50, $n$ = 10 and $k$ = 6. In the sensitivity study, we vary the value of a parameter, but fix the values of the other two parameters. We set the values of parameters within reasonable ranges. Specifically, we vary the value of $k$ from 5 to 7, the value of $m$ from 40 to 60, and the value of $n$ from 8 to 12. We report the evaluation results on the MiniImageNet and Cifar-FS workloads; the results on other workloads are similar, thus omitted here. 
	
The detailed evaluation results have been presented in Table~\ref{table:sensitivity}. It can be observed that the performance of GML only fluctuates marginally ($\leq 0.5\%$ in most cases) as the values of $m$, $n$ and $k$ change. These observations clearly indicate that the performance of GML is very robust w.r.t these parameters. They bode well for their application in real scenarios. 

%% file: 7.Conclusion.tex
\section{Conclusion}\label{sec:Conclusion}
In this paper, we propose a novel solution for few-shot image classification based on the non-i.i.d paradigm of GML. Beginning with only a few labeled samples, it gradually labels unlabeled samples in the increasing order of hardness by iterative factor inference in a factor graph. To facilitate gradual knowledge conveyance, we leverage the existing CNN backbones to extract discriminative image features and model them as monotonous factors in a factor graph. Our extensive experiments on benchmark datasets have validated its efficacy. 

Our research on gradual machine learning is an ongoing endeavor, and there are several avenues for future exploration. First, while this paper targets few-shot image classification, the proposed approach is potentially applicable to other few-shot learning tasks (e.g., object detection, image segmentation). Detailed technical solutions however remain to be investigated. Second, we currently leverage the existing backbones and training procedure to extract deep image features. However, for few-shot gradual learning, deep feature generalization with only a few labeled samples remains its major performance bottleneck. It is very interesting to investigate how to design new backbones for feature extraction that can more effectively support gradual learning.

%% file: template.bbl
\begin{thebibliography}{55}
\providecommand{\natexlab}[1]{#1}
\providecommand{\url}[1]{\texttt{#1}}
\expandafter\ifx\csname urlstyle\endcsname\relax
  \providecommand{\doi}[1]{doi: #1}\else
  \providecommand{\doi}{doi: \begingroup \urlstyle{rm}\Url}\fi

\bibitem[Schmarje et~al.(2021)Schmarje, Santarossa, Schr{\"o}der, and
  Koch]{schmarje2021survey}
Lars Schmarje, Monty Santarossa, Simon-Martin Schr{\"o}der, and Reinhard Koch.
\newblock A survey on semi-, self-and unsupervised learning for image
  classification.
\newblock \emph{IEEE Access}, 9:\penalty0 82146--82168, 2021.

\bibitem[Gwilliam et~al.(2021)Gwilliam, Teuscher, Anderson, and
  Farrell]{2021Fair}
Matthew Gwilliam, Adam Teuscher, Connor Anderson, and Ryan Farrell.
\newblock Fair comparison: Quantifying variance in results for fine-grained
  visual categorization.
\newblock In \emph{Proceedings of the IEEE/CVF Winter Conference on
  Applications of Computer Vision}, pages 3309--3318, 2021.

\bibitem[Feng et~al.(2021)Feng, Zheng, Gao, Chen, Wang, Chen, and
  Wu]{2021Interactive}
Ruiwei Feng, Xiangshang Zheng, Tianxiang Gao, Jintai Chen, Wenzhe Wang, Danny~Z
  Chen, and Jian Wu.
\newblock Interactive few-shot learning: Limited supervision, better medical
  image segmentation.
\newblock \emph{IEEE Transactions on Medical Imaging}, 40\penalty0
  (10):\penalty0 2575--2588, 2021.

\bibitem[Yu et~al.(2020)Yu, Qin, Liu, Wang, and Chen]{yu2020rein}
Fuxun Yu, Zhuwei Qin, Chenchen Liu, Di~Wang, and Xiang Chen.
\newblock Rein the robuts: Robust dnn-based image recognition in autonomous
  driving systems.
\newblock \emph{IEEE Transactions on Computer-Aided Design of Integrated
  Circuits and Systems}, 40\penalty0 (6):\penalty0 1258--1271, 2020.

\bibitem[Wang et~al.(2020{\natexlab{a}})Wang, Yao, Kwok, and
  Ni]{2020Generalizing}
Yaqing Wang, Quanming Yao, James~T Kwok, and Lionel~M Ni.
\newblock Generalizing from a few examples: A survey on few-shot learning.
\newblock \emph{ACM computing surveys (csur)}, 53\penalty0 (3):\penalty0 1--34,
  2020{\natexlab{a}}.

\bibitem[Song et~al.(2022)Song, Wang, Mondal, and Sahoo]{2022A}
Yisheng Song, Ting Wang, Subrota~K Mondal, and Jyoti~Prakash Sahoo.
\newblock A comprehensive survey of few-shot learning: Evolution, applications,
  challenges, and opportunities.
\newblock \emph{arXiv preprint arXiv:2205.06743}, 2022.

\bibitem[Koch et~al.(2015)Koch, Zemel, Salakhutdinov, et~al.]{0Siamese}
Gregory Koch, Richard Zemel, Ruslan Salakhutdinov, et~al.
\newblock Siamese neural networks for one-shot image recognition.
\newblock In \emph{ICML deep learning workshop}, volume~2. Lille, 2015.

\bibitem[Vinyals et~al.(2016)Vinyals, Blundell, Lillicrap, Wierstra,
  et~al.]{2016Matching}
Oriol Vinyals, Charles Blundell, Timothy Lillicrap, Daan Wierstra, et~al.
\newblock Matching networks for one shot learning.
\newblock \emph{Advances in neural information processing systems}, 29, 2016.

\bibitem[Snell et~al.(2017)Snell, Swersky, and Zemel]{2017Prototypical}
Jake Snell, Kevin Swersky, and Richard Zemel.
\newblock Prototypical networks for few-shot learning.
\newblock \emph{Advances in neural information processing systems}, 30, 2017.

\bibitem[Sung et~al.(2018)Sung, Yang, Zhang, Xiang, Torr, and
  Hospedales]{sung2018learning}
Flood Sung, Yongxin Yang, Li~Zhang, Tao Xiang, Philip~HS Torr, and Timothy~M
  Hospedales.
\newblock Learning to compare: Relation network for few-shot learning.
\newblock In \emph{Proceedings of the IEEE conference on computer vision and
  pattern recognition}, pages 1199--1208, 2018.

\bibitem[Ravi and Larochelle(2017)]{ravi2017optimization}
Sachin Ravi and Hugo Larochelle.
\newblock Optimization as a model for few-shot learning.
\newblock In \emph{International conference on learning representations}, 2017.

\bibitem[Finn et~al.(2017)Finn, Abbeel, and Levine]{2017Model}
Chelsea Finn, Pieter Abbeel, and Sergey Levine.
\newblock Model-agnostic meta-learning for fast adaptation of deep networks.
\newblock In \emph{International conference on machine learning}, pages
  1126--1135. PMLR, 2017.

\bibitem[Hou et~al.(2019)Hou, Chen, Shen, Liu, Zhong, Wang, Chen, and
  Li]{hou2019gradual}
Boyi Hou, Qun Chen, Jiquan Shen, Xin Liu, Ping Zhong, Yanyan Wang, Zhaoqiang
  Chen, and Zhanhuai Li.
\newblock Gradual machine learning for entity resolution.
\newblock In \emph{The World Wide Web Conference}, pages 3526--3530, 2019.

\bibitem[Hou et~al.(2022)Hou, Chen, Wang, Nafa, and Li]{HouTKDE}
Boyi Hou, Qun Chen, Yanyan Wang, Youcef Nafa, and Zhanhuai Li.
\newblock Gradual machine learning for entity resolution.
\newblock \emph{IEEE Transactions on Knowledge and Data Engineering},
  34\penalty0 (4):\penalty0 1803--1814, 2022.
\newblock \doi{10.1109/TKDE.2020.3006142}.

\bibitem[Zhong et~al.(2021)Zhong, Li, Chen, and Hou]{zhong2021attention}
Ping Zhong, Zhanhuai Li, Qun Chen, and Boyi Hou.
\newblock Attention-enhanced gradual machine learning for entity resolution.
\newblock \emph{IEEE Intelligent Systems}, 36\penalty0 (6):\penalty0 71--79,
  2021.

\bibitem[Wang et~al.(2020{\natexlab{b}})Wang, Chen, Shen, Hou, and
  Li]{2020Aspect}
Y.~Wang, Q.~Chen, J.~Shen, B.~Hou, and Z.~Li.
\newblock Aspect-level sentiment analysis based on gradual machine learning.
\newblock \emph{Knowledge-Based Systems}, 212:\penalty0 106509,
  2020{\natexlab{b}}.

\bibitem[Ahmed et~al.(2021)Ahmed, Chen, Wang, Nafa, Li, and Duan]{ahmed2021dnn}
Murtadha Ahmed, Qun Chen, Yanyan Wang, Youcef Nafa, Zhanhuai Li, and Tianyi
  Duan.
\newblock Dnn-driven gradual machine learning for aspect-term sentiment
  analysis.
\newblock In \emph{Findings of the Association for Computational Linguistics:
  ACL-IJCNLP 2021}, pages 488--497, 2021.

\bibitem[Wang et~al.(2023)Wang, Chen, Ahmed, Chen, Su, Pan, and
  Li]{wang2023supervised}
Yanyan Wang, Qun Chen, Murtadha~HM Ahmed, Zhaoqiang Chen, Jing Su, Wei Pan, and
  Zhanhuai Li.
\newblock Supervised gradual machine learning for aspect-term sentiment
  analysis.
\newblock \emph{Transactions of the Association for Computational Linguistics},
  11:\penalty0 723--739, 2023.

\bibitem[Afrasiyabi et~al.(2022)Afrasiyabi, Larochelle, Lalonde, and
  Gagn{\'e}]{afrasiyabi2022matching}
Arman Afrasiyabi, Hugo Larochelle, Jean-Fran{\c{c}}ois Lalonde, and Christian
  Gagn{\'e}.
\newblock Matching feature sets for few-shot image classification.
\newblock In \emph{Proceedings of the IEEE/CVF Conference on Computer Vision
  and Pattern Recognition}, pages 9014--9024, 2022.

\bibitem[Zhao et~al.(2022)Zhao, Liu, Cao, Lian, and He]{zhao2022self}
Zhineng Zhao, Qifan Liu, Wenming Cao, Deliang Lian, and Zhihai He.
\newblock Self-guided information for few-shot classification.
\newblock \emph{Pattern Recognition}, 131:\penalty0 108880, 2022.

\bibitem[He et~al.(2016)He, Zhang, Ren, and Sun]{2016Deep}
Kaiming He, Xiangyu Zhang, Shaoqing Ren, and Jian Sun.
\newblock Deep residual learning for image recognition.
\newblock In \emph{Proceedings of the IEEE conference on computer vision and
  pattern recognition}, pages 770--778, 2016.

\bibitem[Zagoruyko and Komodakis(2016)]{2016Wide}
Sergey Zagoruyko and Nikos Komodakis.
\newblock Wide residual networks.
\newblock \emph{arXiv preprint arXiv:1605.07146}, 2016.

\bibitem[Bendou et~al.(2022)Bendou, Hu, Lafargue, Lioi, Pasdeloup, Pateux, and
  Gripon]{bendou2022easy}
Yassir Bendou, Yuqing Hu, Raphael Lafargue, Giulia Lioi, Bastien Pasdeloup,
  St{\'e}phane Pateux, and Vincent Gripon.
\newblock Easy: Ensemble augmented-shot y-shaped learning: State-of-the-art
  few-shot classification with simple ingredients.
\newblock \emph{arXiv preprint arXiv:2201.09699}, 2022.

\bibitem[Zhang et~al.(2020)Zhang, Cai, Lin, and Shen]{zhang2020deepemd}
Chi Zhang, Yujun Cai, Guosheng Lin, and Chunhua Shen.
\newblock Deepemd: Few-shot image classification with differentiable earth
  mover's distance and structured classifiers.
\newblock In \emph{Proceedings of the IEEE/CVF conference on computer vision
  and pattern recognition}, pages 12203--12213, 2020.

\bibitem[Wertheimer et~al.(2021)Wertheimer, Tang, and
  Hariharan]{wertheimer2021few}
Davis Wertheimer, Luming Tang, and Bharath Hariharan.
\newblock Few-shot classification with feature map reconstruction networks.
\newblock In \emph{Proceedings of the IEEE/CVF Conference on Computer Vision
  and Pattern Recognition}, pages 8012--8021, 2021.

\bibitem[Chen et~al.(2019)Chen, Liu, Kira, Wang, and Huang]{chen2019closer}
Wei-Yu Chen, Yen-Cheng Liu, Zsolt Kira, Yu-Chiang~Frank Wang, and Jia-Bin
  Huang.
\newblock A closer look at few-shot classification.
\newblock \emph{arXiv preprint arXiv:1904.04232}, 2019.

\bibitem[Zhu and Koniusz(2022)]{zhu2022ease}
Hao Zhu and Piotr Koniusz.
\newblock Ease: Unsupervised discriminant subspace learning for transductive
  few-shot learning.
\newblock In \emph{Proceedings of the IEEE/CVF Conference on Computer Vision
  and Pattern Recognition}, pages 9078--9088, 2022.

\bibitem[Hu et~al.(2022)Hu, Pateux, and Gripon]{hu2022squeezing}
Yuqing Hu, St{\'e}phane Pateux, and Vincent Gripon.
\newblock Squeezing backbone feature distributions to the max for efficient
  few-shot learning.
\newblock \emph{Algorithms}, 15\penalty0 (5):\penalty0 147, 2022.

\bibitem[Hu et~al.(2021)Hu, Gripon, and Pateux]{hu2021leveraging}
Yuqing Hu, Vincent Gripon, and St{\'e}phane Pateux.
\newblock Leveraging the feature distribution in transfer-based few-shot
  learning.
\newblock In \emph{Artificial Neural Networks and Machine Learning--ICANN 2021:
  30th International Conference on Artificial Neural Networks, Bratislava,
  Slovakia, September 14--17, 2021, Proceedings, Part II 30}, pages 487--499.
  Springer, 2021.

\bibitem[Jiang et~al.(2020)Jiang, Huang, Geng, and Deng]{jiang2020multi}
Wen Jiang, Kai Huang, Jie Geng, and Xinyang Deng.
\newblock Multi-scale metric learning for few-shot learning.
\newblock \emph{IEEE Transactions on Circuits and Systems for Video
  Technology}, 31\penalty0 (3):\penalty0 1091--1102, 2020.

\bibitem[Pan and Yang(2009)]{pan2009survey}
Sinno~Jialin Pan and Qiang Yang.
\newblock A survey on transfer learning.
\newblock \emph{IEEE Transactions on knowledge and data engineering},
  22\penalty0 (10):\penalty0 1345--1359, 2009.

\bibitem[Verma et~al.(2019)Verma, Lamb, Beckham, Najafi, Mitliagkas, Lopez-Paz,
  and Bengio]{verma2019manifold}
Vikas Verma, Alex Lamb, Christopher Beckham, Amir Najafi, Ioannis Mitliagkas,
  David Lopez-Paz, and Yoshua Bengio.
\newblock Manifold mixup: Better representations by interpolating hidden
  states.
\newblock In \emph{International conference on machine learning}, pages
  6438--6447. PMLR, 2019.

\bibitem[Mangla et~al.(2020)Mangla, Kumari, Sinha, Singh, Krishnamurthy, and
  Balasubramanian]{mangla2020charting}
Puneet Mangla, Nupur Kumari, Abhishek Sinha, Mayank Singh, Balaji
  Krishnamurthy, and Vineeth~N Balasubramanian.
\newblock Charting the right manifold: Manifold mixup for few-shot learning.
\newblock In \emph{Proceedings of the IEEE/CVF winter conference on
  applications of computer vision}, pages 2218--2227, 2020.

\bibitem[Wang et~al.(2019)Wang, Chao, Weinberger, and van~der
  Maaten]{wang2019simpleshot}
Yan Wang, Wei-Lun Chao, Kilian~Q Weinberger, and Laurens van~der Maaten.
\newblock Simpleshot: Revisiting nearest-neighbor classification for few-shot
  learning.
\newblock \emph{arXiv preprint arXiv:1911.04623}, 2019.

\bibitem[Tian et~al.(2020)Tian, Wang, Krishnan, Tenenbaum, and
  Isola]{tian2020rethinking}
Yonglong Tian, Yue Wang, Dilip Krishnan, Joshua~B Tenenbaum, and Phillip Isola.
\newblock Rethinking few-shot image classification: a good embedding is all you
  need?
\newblock In \emph{Computer Vision--ECCV 2020: 16th European Conference,
  Glasgow, UK, August 23--28, 2020, Proceedings, Part XIV 16}, pages 266--282.
  Springer, 2020.

\bibitem[Kim et~al.(2019)Kim, Kim, Kim, and Yoo]{kim2019edge}
Jongmin Kim, Taesup Kim, Sungwoong Kim, and Chang~D Yoo.
\newblock Edge-labeling graph neural network for few-shot learning.
\newblock In \emph{Proceedings of the IEEE/CVF conference on computer vision
  and pattern recognition}, pages 11--20, 2019.

\bibitem[Yang et~al.(2020)Yang, Li, Zhang, Zhou, Zhou, and Liu]{yang2020dpgn}
Ling Yang, Liangliang Li, Zilun Zhang, Xinyu Zhou, Erjin Zhou, and Yu~Liu.
\newblock Dpgn: Distribution propagation graph network for few-shot learning.
\newblock In \emph{Proceedings of the IEEE/CVF conference on computer vision
  and pattern recognition}, pages 13390--13399, 2020.

\bibitem[Ma et~al.(2020)Ma, Bai, An, Liu, Liu, Zhen, and
  Liu]{ma2020transductive}
Yuqing Ma, Shihao Bai, Shan An, Wei Liu, Aishan Liu, Xiantong Zhen, and
  Xianglong Liu.
\newblock Transductive relation-propagation network for few-shot learning.
\newblock In \emph{IJCAI}, volume~20, pages 804--810, 2020.

\bibitem[Zhu and Koniusz(2023)]{zhu2023transductive}
Hao Zhu and Piotr Koniusz.
\newblock Transductive few-shot learning with prototype-based label propagation
  by iterative graph refinement.
\newblock In \emph{Proceedings of the IEEE/CVF Conference on Computer Vision
  and Pattern Recognition}, pages 23996--24006, 2023.

\bibitem[Liu et~al.(2020)Liu, Song, and Qin]{liu2020prototype}
Jinlu Liu, Liang Song, and Yongqiang Qin.
\newblock Prototype rectification for few-shot learning.
\newblock In \emph{Computer Vision--ECCV 2020: 16th European Conference,
  Glasgow, UK, August 23--28, 2020, Proceedings, Part I 16}, pages 741--756.
  Springer, 2020.

\bibitem[Chobola et~al.(2021)Chobola, Va{\v{s}}ata, and
  Kord{\'\i}k]{chobola2021transfer}
Tom{\'a}{\v{s}} Chobola, Daniel Va{\v{s}}ata, and Pavel Kord{\'\i}k.
\newblock Transfer learning based few-shot classification using optimal
  transport mapping from preprocessed latent space of backbone neural network.
\newblock In \emph{AAAI Workshop on Meta-Learning and MetaDL Challenge}, pages
  29--37. PMLR, 2021.

\bibitem[Huang et~al.(2019)Huang, Larochelle, and Lacoste-Julien]{huang2019few}
Gabriel Huang, Hugo Larochelle, and Simon Lacoste-Julien.
\newblock Are few-shot learning benchmarks too simple?
\newblock 2019.

\bibitem[Shalam and Korman(2022)]{shalam2022self}
Daniel Shalam and Simon Korman.
\newblock The self-optimal-transport feature transform.
\newblock \emph{arXiv e-prints}, pages arXiv--2204, 2022.

\bibitem[Wang et~al.(2021)Wang, Chen, Shen, Hou, Ahmed, and Li]{wang2021aspect}
Yanyan Wang, Qun Chen, Jiquan Shen, Boyi Hou, Murtadha Ahmed, and Zhanhuai Li.
\newblock Aspect-level sentiment analysis based on gradual machine learning.
\newblock \emph{Knowledge-Based Systems}, 212:\penalty0 106509, 2021.

\bibitem[Hou et~al.(2018)Hou, Chen, Chen, Nafa, and Li]{hou2018r}
Boyi Hou, Qun Chen, Zhaoqiang Chen, Youcef Nafa, and Zhanhuai Li.
\newblock R-humo: A risk-aware human-machine cooperation framework for entity
  resolution with quality guarantees.
\newblock \emph{IEEE Transactions on Knowledge and Data Engineering},
  32\penalty0 (2):\penalty0 347--359, 2018.

\bibitem[Ren et~al.(2018)Ren, Triantafillou, Ravi, Snell, Swersky, Tenenbaum,
  Larochelle, and Zemel]{ren2018meta}
Mengye Ren, Eleni Triantafillou, Sachin Ravi, Jake Snell, Kevin Swersky,
  Joshua~B Tenenbaum, Hugo Larochelle, and Richard~S Zemel.
\newblock Meta-learning for semi-supervised few-shot classification.
\newblock \emph{arXiv preprint arXiv:1803.00676}, 2018.

\bibitem[Bertinetto et~al.(2018)Bertinetto, Henriques, Torr, and
  Vedaldi]{bertinetto2018meta}
Luca Bertinetto, Joao~F Henriques, Philip~HS Torr, and Andrea Vedaldi.
\newblock Meta-learning with differentiable closed-form solvers.
\newblock \emph{arXiv preprint arXiv:1805.08136}, 2018.

\bibitem[Wah et~al.(2011)Wah, Branson, Welinder, Perona, and
  Belongie]{wah2011caltech}
Catherine Wah, Steve Branson, Peter Welinder, Pietro Perona, and Serge
  Belongie.
\newblock The caltech-ucsd birds-200-2011 dataset.
\newblock 2011.

\bibitem[Zhang et~al.(2021)Zhang, Meng, Gouk, and Hospedales]{zhang2021shallow}
Xueting Zhang, Debin Meng, Henry Gouk, and Timothy~M Hospedales.
\newblock Shallow bayesian meta learning for real-world few-shot recognition.
\newblock In \emph{Proceedings of the IEEE/CVF International Conference on
  Computer Vision}, pages 651--660, 2021.

\bibitem[Ziko et~al.(2020)Ziko, Dolz, Granger, and Ayed]{ziko2020laplacian}
Imtiaz Ziko, Jose Dolz, Eric Granger, and Ismail~Ben Ayed.
\newblock Laplacian regularized few-shot learning.
\newblock In \emph{International conference on machine learning}, pages
  11660--11670. PMLR, 2020.

\bibitem[Liu et~al.(2023)Liu, Cao, and He]{liu2023cycle}
Qifan Liu, Wenming Cao, and Zhihai He.
\newblock Cycle optimization metric learning for few-shot classification.
\newblock \emph{Pattern Recognition}, page 109468, 2023.

\bibitem[Kye et~al.(2020)Kye, Lee, Kim, and Hwang]{kye2020meta}
Seong~Min Kye, Hae~Beom Lee, Hoirin Kim, and Sung~Ju Hwang.
\newblock Meta-learned confidence for transductive few-shot learning.
\newblock 2020.

\bibitem[Bateni et~al.(2022)Bateni, Barber, Van~de Meent, and
  Wood]{bateni2022enhancing}
Peyman Bateni, Jarred Barber, Jan-Willem Van~de Meent, and Frank Wood.
\newblock Enhancing few-shot image classification with unlabelled examples.
\newblock In \emph{Proceedings of the IEEE/CVF Winter Conference on
  Applications of Computer Vision}, pages 2796--2805, 2022.

\bibitem[Kang et~al.(2021)Kang, Kwon, Min, and Cho]{2021Relational}
D.~Kang, H.~Kwon, J.~Min, and M.~Cho.
\newblock Relational embedding for few-shot classification.
\newblock 2021.

\bibitem[Lazarou et~al.(2021)Lazarou, Stathaki, and
  Avrithis]{Lazarou_2021_ICCV}
Michalis Lazarou, Tania Stathaki, and Yannis Avrithis.
\newblock Iterative label cleaning for transductive and semi-supervised
  few-shot learning.
\newblock In \emph{Proceedings of the IEEE/CVF International Conference on
  Computer Vision (ICCV)}, pages 8751--8760, October 2021.

\end{thebibliography}
